\documentclass{article}

\usepackage{arxiv}
\usepackage[utf8]{inputenc}
\usepackage[T1]{fontenc} 
\usepackage{hyperref}      
\usepackage{url}          
\usepackage{booktabs}      
\usepackage{amsfonts}   
\usepackage{amsmath}
\usepackage{xcolor}
\usepackage{nicefrac}      
\usepackage{microtype}  
\usepackage{lipsum}		
\usepackage{graphicx}
\usepackage{doi}
\usepackage{caption}
\usepackage{subcaption}

\title{GP-KAN: Gaussian Process Kolmogorov-Arnold Networks}

\author{
	{\hspace{1mm}Andrew Siyuan Chen}\thanks{Portion of this research was done while author was a student at University of Cambridge, Engineering Department}\thanks{This paper currently under review at IEEE. Copyright notice:  “© © 2024 IEEE. Personal use of this material is permitted. Permission from IEEE must be obtained for all other uses, in any current or future media, including reprinting/republishing this material for advertising or promotional purposes, creating new collective works, for resale or redistribution to servers or lists, or reuse of any copyrighted component of this work in other works.”} \\
    sc2178@cantab.ac.uk
}

\date{}

\renewcommand{\headeright}{}
\renewcommand{\undertitle}{}

\begin{document}
\maketitle

\begin{abstract}
    In this paper, we introduce a probabilistic extension to Kolmogorov Arnold Networks (KANs) by incorporating Gaussian Process (GP) as non-linear neurons, which we refer to as GP-KAN. A fully analytical approach to handling the output distribution of one GP as an input to another GP is achieved by considering the function inner product of a GP function sample with the input distribution. These GP neurons exhibit robust non-linear modelling capabilities while using few parameters and can be easily and fully integrated in a feed-forward network structure. They provide inherent uncertainty estimates to the model prediction and can be trained directly on the log-likelihood objective function, without needing variational lower bounds or approximations. In the context of MNIST classification, a model based on GP-KAN of 80 thousand parameters achieved 98.5\% prediction accuracy, compared to current state-of-the-art models with 1.5 million parameters.
\end{abstract}


\section{Introduction}
Neural Networks have traditionally relied on linear neurons \cite{ANN} connected in a feed-forward graph layout. The depth of these models has grown greatly over the years, to capture highly non-linear features in the context of increasingly diverse applications. These deep networks also require well-chosen activation functions (e.g. Sigmoid, ReLU) that can stabilise training, keep gradients from diminishing or exploding, and transform the intermediate features to appropriate representations in a forward pass. Much of a model's architecture revolves around choosing sufficient depth and width to model the complicated non-linear relationships between a given input and a desired output.

Recently, non-linear neurons called Kolmogorov Arnold Networks (KAN) \cite{liu2024kan} have been proposed. Instead of using deep feed-forward structures to capture non-linearity, KAN offers an alternative representation framework rooted in mathematical theory where each neuron can directly model complex non-linearity. Still in the early exploration stage of its potential, KAN variants have shown application in Symbolic Regression and Image Classification \cite{ConvolutionalKAN}. The non-linear neurons offer the potential to reduce model parameter count while maintaining or even improving modelling capabilities.

In this paper, a probabilistic extension to Kolmogorov Arnold Network architecture is proposed. The B-Spline used in the non-linear neurons of KAN is replaced by a Gaussian Process (GP), creating a form of Deep GP model. GPs describe a probability distribution over functions and have been interpreted as the theoretical limit of infinitely-wide Deep Neural Networks \cite{InfiniteWidthNNasGP,lee2018deep}. GPs have shown strong representational capabilities as non-parametric models conditioned on a training dataset. Research has been done on arranging GPs in a feed-forward layout to produce deep GP models \cite{deepGP} to improve their representation capabilities. Uncertainty propagation in Deep GP models is usually an issue due to the intractable posterior when there are multiple GP layers in the model. Some typical solutions rely on variational lower bounds \cite{deepGP} and approximate expectation propagation \cite{DeepGPapproximateEP}, but these methods have their own limitations. For example, optimising a deep GP model on the variational lower bound of the posterior can be ineffective if the lower bound expression turns out to be far from the true posterior.

To overcome the above-mentioned issue, this work proposes "collapsing" the input prior distribution to a GP by considering the inner product of a GP function sample with the input distribution, producing an output Gaussian Distribution. Inspired by KANs, only one-dimensional GPs are used to simplify the analysis. The Gaussian nature is maintained on the input and output to each layer, allowing exact analytical uncertainty propagation. Analogous to KANs, the only binary operation is the addition of Gaussian random variables, thus allowing the new GP model to be both deep and wide. This approach enables model structures similar to those already familiar to the Machine Learning community, such as Fully-Connected and Convolutional layers, with the additional benefit of more expressive non-linear neurons and inherent uncertainty modelling. As an example, a simple model setup trained  on the MNIST dataset, with only 80,000 parameters, achieved a test accuracy of 98.5\%, in contrast to current state-of-the-art models that uses 1.5 million parameters.

\section{Background}
\subsection{Kolmogorov-Arnold Networks}
Kolmogorov-Arnold Networks (KANs) \cite{liu2024kan} are class of networks inspired by the Kolmogorov-Arnold Representation Theorem \cite{AndreyKolmogorov,VladimirArnold}. This theorem states that any bounded multivariate continuous function $f$ can be represented as a composition of univariate functions $\phi$ and the binary operator of addition. The original formulation of the theorem examines a composition of 2 layers:
\begin{equation}
    f(x_1, x_2, ... x_n) = \sum_{q=1}^{2n}\phi_{q,n+1}\left(\sum_{p=1}^{n}\phi_{q,p}(x_p)\right)
    \label{eq:Kolmogorov-Arnold-Theorem}
\end{equation}
However, the theorem provides no guarantees on the nature of $\phi$, which can be non-smooth and possess characteristics that hinder learning process. KANs generalizes the concept to arbitrary depths and widths, such that $\phi$ becomes learnable via traditional machine learning methods such as back propagation. The KANs proposed in \cite{liu2024kan} use B-spline to approximate $\phi$. They also demonstrate that with the appropriate choices of basis functions, KAN is able to recover the exact symbolic formulation of various multivariate functions, and sometimes even discover alternative formulations. Further work \cite{li2024kolmogorovarnold} expanding on the use of Radial Basis Functions (RBF) has successfully applied KANs to machine learning tasks such as MNIST image classification \cite{MNIST}. It becomes evident that the choice of basis functions greatly impacts the modelling capabilities of KANs.

\subsection{Gaussian Process}
Gaussian Process (GP) \cite{GaussianProcess} describes the probability distribution over functions by treating functions as a collection of Random Variables (r.v.) where any finite subset has a joint Gaussian Distribution. A GP is fully specified by its mean function $m(\cdot)$ and covariance function $k(\cdot, \cdot)$.
\begin{equation}
    \begin{aligned}
        f                                    & \sim\mathcal{GP}(m,k)
        \\
        \text{where} \quad m(x) & =\mathbb{E} [f(x)]
        \\
        k(x,x')    & =\text{Covar}[f(x), f(x')]
    \end{aligned}
    \label{eq:GP_def}
\end{equation}
Given a set of known noisy function values $\boldsymbol{h}$ at locations $\boldsymbol{z}$, GP prediction at a new location $x$ follows Equation \ref{eq:GP_pred}. In GP regression, $(\boldsymbol{h}, \boldsymbol{z})$ corresponds to known data labels and inputs, with label noise $\sigma_n^2$ , and the training objective of maximising their likelihood (or equivalently, the log-likelihood). A key drawback of Gaussian Process is the \textbf{curse of dimensionality}, due to the $\mathcal{O}(n^3)$ complexity of inverting $K_{hh}$ in Equation \ref{eq:GP_pred} for $n$ data points and the need for large $(\boldsymbol{h}, \boldsymbol{z})$ to map the input space for high-dimensional inputs.
\begin{equation}
    \begin{aligned}
        p(f|\boldsymbol{h},\boldsymbol{z},x)     & =\mathcal{N}(\mu, \Sigma)           \\
        \text{where}\quad\quad\mu & =m(x) + \boldsymbol{k}_{xh}(K_{hh}+\sigma_n^2 I)^{-1}\boldsymbol{h} \\
        \Sigma                    & =k(x, x)-\boldsymbol{k}_{xh}(K_{hh}+\sigma_n^2 I)^{-1}\boldsymbol{k}_{hx}        \\
        \boldsymbol{k}_{xh}^T                  & =\boldsymbol{k}_{hx}=\begin{bmatrix}
                                                k(z_1, x) \\
                                                k(z_2, x) \\
                                                \vdots
                                            \end{bmatrix}                  \\
        K_{hh}                    & =\begin{bmatrix}
                                         k(z_1,z_1) & k(z_1, z_2) & \hdots \\
                                         k(z_2,z_1) &             &        \\
                                         \vdots     &             &
                                     \end{bmatrix}
    \end{aligned}
    \label{eq:GP_pred}
\end{equation}

\section{GP-KAN}
GPs and KANs have the potential to be good complements. KANs only considers the usage of univariate functions $\phi$, avoiding GPs' \textbf{curse of dimensionality}. For a given input $x$, and known data $(\boldsymbol{h}, \boldsymbol{z})$, the predictive GP defines the distribution of an output Gaussian r.v., $p(f|\boldsymbol{h},\boldsymbol{z},x)$. The set of Gaussian random variables is closed under the binary operation of addition, the only multivariate operation in KANs. On the other hand, KANs can benefit from good choices of basis functions for $\phi$, and it can be shown that the \textit{squared exponential} covariance function (Equation \ref{eq:covar_se}) corresponds to a Bayesian Linear Regression Model of infinite number of RBFs (Sect 4.3 \& Eq 4.13 in \cite{GaussianProcess}).
\begin{equation}
    k(x,x')=s^2\exp\left(-\frac{(x-x')^2}{2l^2}\right)=s^2l\sqrt{2\pi}\mathcal{N}(x|x',l^2)
    \label{eq:covar_se}
\end{equation}
Almost immediately, however, one runs into the problem of uncertainty propagation. For an input value that is Gaussian distributed $p(x)$, the output posterior distribution $p(f|\boldsymbol{h},\boldsymbol{z})=\int p(f|\boldsymbol{h},\boldsymbol{z},x)p(x)dx$ is often intractable and non-Gaussian \cite{deepGP, GPLVM}. This presents a significant obstacle in using multi-layer GP networks whilst maintaining a fully Bayesian treatment. One solution is Monte-Carlo estimation, which suffers from steep computational cost and poor estimation in high-dimensional problem sets. Other approaches \cite{deepGP} rely on variational methods to optimise a tractable strict lower bound to the objective function, but the model can struggle to learn the desired features if the chosen lower bound is far from the true posterior.

In this work, we decide to focus on maintaining the following characteristics of the deep GP model:
\begin{enumerate}
    \item Elements propagated from layer to layer are Gaussian-distributed and independent
    \item All GPs within the model are univariate GPs
\end{enumerate}
These help to simplify analysis and produce a tractable exact objective function, while avoiding the need to track covariance between intermediate elements of the model, keeping the computational complexity to $O(n)$ for $n$ intermediate elements.

\subsection{Single GP neuron}
First, for a linear transformation $W$, vector $\boldsymbol{b}$, mean $\mu$ and covariance matrices $\Sigma_1$ and $\Sigma_2$, the equality holds:
\begin{equation}
    \int\mathcal{N}(\boldsymbol{r}|W\boldsymbol{x}+\boldsymbol{b},\Sigma_2)\mathcal{N}(\boldsymbol{x}|\mu,\Sigma_1)d\boldsymbol{x}=\mathcal{N}(\boldsymbol{r}|W\mu+\boldsymbol{b},W\Sigma_1W^T+\Sigma_2)
    \label{eq:int_gaussian}
\end{equation}

Equation \ref{eq:int_gaussian} can be proven by considering Gaussian random variables $\Tilde{\boldsymbol{x}}\sim\mathcal{N}(\boldsymbol{x}|\mu,\Sigma_1)$ and $\Tilde{\boldsymbol{e}}\sim\mathcal{N}(\boldsymbol{e}|\boldsymbol{b},\Sigma_2)$. Then the Gaussian random variable $\Tilde{\boldsymbol{r}}=(W\Tilde{\boldsymbol{x}}+\Tilde{\boldsymbol{e}})\sim\mathcal{N}(\boldsymbol{y}|W\mu+\boldsymbol{b},W\Sigma_1W^T+\Sigma_2)$. Marginalizing the distribution for $\Tilde{\boldsymbol{r}}$ gives $p(\boldsymbol{r})=\int p(\boldsymbol{r}|\boldsymbol{x})p(\boldsymbol{x})d\boldsymbol{x}$, where $p(\boldsymbol{x})=\mathcal{N}(\boldsymbol{x}|\mu,\Sigma_1)$ and $p(\boldsymbol{r}|\boldsymbol{x})=\mathcal{N}(\boldsymbol{r}|W\boldsymbol{x}+\boldsymbol{b},\Sigma_2)$, thus yielding Equation \ref{eq:int_gaussian}. This equation is also available in a more general form \cite{MatrixCookBook}.

Next, consider the following setting up the random variable $\tilde{y}$ such that
\begin{equation}
\begin{aligned}
	\tilde{y}&=\langle \tilde{f}, p(x)\rangle=\int \tilde{f}(x) p(x)dx \\
    \text{ for }p(x)&=\mathcal{N}(x|\mu_x,\sigma_x^2) \\
    \tilde{f}&\sim\mathcal{GP}(m(\cdot),k(\cdot,\cdot)|\boldsymbol{h},\boldsymbol{z})
    \label{eq:reformulation}
\end{aligned}
\end{equation}
where $\langle\cdot,\cdot\rangle$ is the function inner product, and $\tilde{f}$ is a function sample drawn from the GP conditioned on data points $(\boldsymbol{h},\boldsymbol{z})$. In the rest of the paper, we will use a mean function $m(\cdot)=0$ and a covariance function $k(\cdot, \cdot)$ that follows Equation \ref{eq:covar_se}. Equation \ref{eq:reformulation} can be viewed as a summation over infinite number of Gaussian-distributed $\tilde{f}(x)$, each scaled by $p(x)$. Thus, $\tilde{y}$ is also Gaussian-distributed, preserving the Gaussian nature on both input and output. The mean of $\tilde{y}$ can be obtained analytically by using Equations \ref{eq:GP_pred} and \ref{eq:int_gaussian}:
\begin{equation}
    \begin{aligned}
        \mathbb{E}\left[\tilde{y}\right]&=\int p(x)\mathbb{E}\left[\tilde{f}(x)\right]dx \\
        &=\int p(x)\left(m(x)+\boldsymbol{k}_{xh}K_{hh}^{-1}\boldsymbol{h}\right)dx \\
        &=\begin{bmatrix}
            \int p(x)k(x,z_1)dx \\
            \int p(x)k(x,z_2)dx \\
            \vdots
        \end{bmatrix}^T K_{hh}^{-1}\boldsymbol{h} \\
        &=\sqrt{2\pi}s^2l\begin{bmatrix}
            \mathcal{N}(\mu_x|z_1,\sigma_x^2+l^2) \\
            \mathcal{N}(\mu_x|z_2,\sigma_x^2+l^2) \\
            \vdots
        \end{bmatrix}^TK_{hh}^{-1}\boldsymbol{h} \\
        &=\sqrt{2\pi}s^2l\boldsymbol{q}_{xh}K_{hh}^{-1}\boldsymbol{h} \\
        \text{where }\boldsymbol{q}_{xh}&=\boldsymbol{q}_{hx}^T=\begin{bmatrix}
            \mathcal{N}(\mu_x|z_1,\sigma_x^2+l^2) \\
            \mathcal{N}(\mu_x|z_2,\sigma_x^2+l^2) \\
            \vdots
        \end{bmatrix}^T
    \end{aligned}
\end{equation}
Similarly, the variance can also be obtained:
\begin{equation}
    \begin{aligned}
        \text{Var}\left[\tilde{y}\right]&=\int\int p(x)\text{Covar}\left[\tilde{f}(x),\tilde{f}(x')\right]p(x')dxdx' \\
        &=\textcolor{red}{\int\int p(x)k(x,x')p(x')dxdx'}
        -\textcolor{blue}{\int\int p(x)\boldsymbol{k}_{xh}K_{hh}^{-1}\boldsymbol{k}_{hx'}p(x')dxdx'}
    \label{eq:incomplete_var}
    \end{aligned}
\end{equation}
where the red term
\begin{equation}
    \begin{aligned}
        \textcolor{red}{\int\int p(x)k(x,x')p(x')dxdx'}&=\int\left(\int p(x)k(x,x')dx\right)p(x')dx \\
        &=\int \sqrt{2\pi}s^2l\mathcal{N}(\mu_x|x',l^2 + \sigma_x^2)p(x')dx' \\
        &=\sqrt{2\pi}s^2l\mathcal{N}(\mu_x|\mu_x,l^2+2\sigma_x^2) \\
        &=\frac{s^2l}{\sqrt{l^2+2\sigma_x^2}}
    \end{aligned}
\end{equation}
and the blue term
\begin{equation}
    \begin{aligned}
        \textcolor{blue}{\int\int p(x)\boldsymbol{k}_{xh}K_{hh}^{-1}\boldsymbol{k}_{hx'}p(x')dxdx'}
        &=\int\left(\int p(x)\boldsymbol{k}_{xh}dx\right)K_{hh}^{-1}\boldsymbol{k}_{hx'}p(x')dx' \\
        &=\int \sqrt{2\pi}s^2l\boldsymbol{q}_{xh}K_{hh}^{-1}\boldsymbol{k}_{hx'}p(x')dx' \\
        &=\sqrt{2\pi}s^2l\boldsymbol{q}_{xh}K_{hh}^{-1} \int \boldsymbol{k}_{hx'}p(x')dx' \\
        &=2\pi s^4 l^2 \boldsymbol{q}_{xh}K_{hh}^{-1}\boldsymbol{q}_{hx}
    \end{aligned}
\end{equation}
Therefore, the overall analytical mean and variance for $\tilde{y}$ can be expressed as follows:
\begin{equation}
    \begin{aligned}
    \mathbb{E}\left[\tilde{y}\right]&=\sqrt{2\pi}s^2l\boldsymbol{q}_{xh}K_{hh}^{-1}\boldsymbol{h} \\
    \text{Var}\left[\tilde{y}\right]&=\frac{s^2l}{\sqrt{l^2+2\sigma_x^2}} - 2\pi s^4 l^2 \boldsymbol{q}_{xh}K_{hh}^{-1}\boldsymbol{q}_{hx}
    \end{aligned}
    \label{eq:gpneuron}
\end{equation}
As a sanity check, in the case of a deterministic input $x$, one has the input variance $\sigma_x^2=0$, and the expression for $p(y)$ returns to the original GP posterior in Equation \ref{eq:GP_pred}.

With this approach, the input prior distribution is effectively "collapsed", by considering not just one sample drawn from the input distribution. Instead, it is equivalent to drawing infinitely many samples from the input distribution, fully covering the input space, and using them as input locations on which the function sample can be drawn from the induced GP. Empirically, this corresponds to the below steps, for a large $N$:
\begin{enumerate}
    \item Draw $\boldsymbol{x}=[x_1\: x_2\: ... \: x_N]^T$ from $x_i\sim \mathcal{N}(\mu_x,\sigma_x^2)$.
    \item Compute the GP predictive posterior $p(\boldsymbol{f}|\boldsymbol{h},\boldsymbol{z})$ at locations $\boldsymbol{x}$, for $f\sim\mathcal{GP}(m(\cdot),k(\cdot,\cdot))$.
    \item Draw a sample $\boldsymbol{f}\sim p(\boldsymbol{f}|\boldsymbol{h},\boldsymbol{z})$ and take $y=\frac{1}{N}\sum_{i=1}^{N}f_i$.
\end{enumerate}

\begin{figure}[t]
    \centering
    \includegraphics[width=0.6\columnwidth]{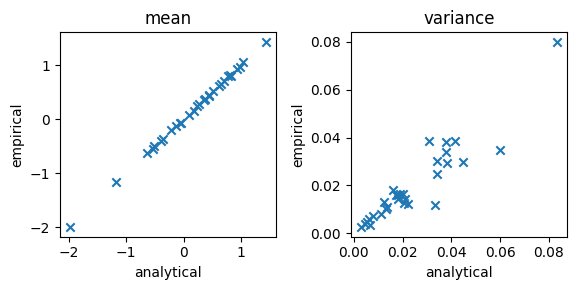}
    \caption{Empirical vs analytical mean and variance values, repeated for different $\mu_x,\sigma_x^2$}
    \label{fig:GP_sim}
\end{figure}

Figure \ref{fig:GP_sim} shows the comparison between the empirical mean and variance values obtained from the above steps and the analytical mean and variance values obtained from Equation \ref{eq:gpneuron}. Since $\tilde{y}$ is the inner product between $p(x)$ and function sample $\tilde{f}$ from the GP conditioned on data $(\boldsymbol{h},\boldsymbol{z})$, $\tilde{y}$ is independent of $\tilde{x}$ and conditioned on internal GP "inducing points" $(\boldsymbol{h},\boldsymbol{z})$. The term "inducing points" is typically used in the framework of Sparse GP \cite{inducingGP,variationalInducingGP}, where a sparse GP conditioned on a smaller inducing set approximately captures the features that GP conditioned on the full dataset would have. Similarly, the inducing points $(\boldsymbol{h},\boldsymbol{z})$ are adapted over the training process to induce a GP that captures the appropriate features according to an objective function.

Linear Neurons \cite{ANN} and non-linear KANs \cite{liu2024kan} both take an input value and produce an output value. The GP neuron proposed here takes a Gaussian distribution as input, and outputs another Gaussian distribution. The trainable parameters of the GP neuron would be the GP inducing points $(\boldsymbol{h}, \boldsymbol{z})$, inducing noise $\sigma_n^2$ and the \textit{squared exponential} covariance function parameters $l, s$.

\subsection{Fully-Connected GP Layer (FCGP)}

\begin{figure}[t]
    \centering
    \begin{subfigure}[b]{0.95\columnwidth}
        \includegraphics[width=\textwidth]{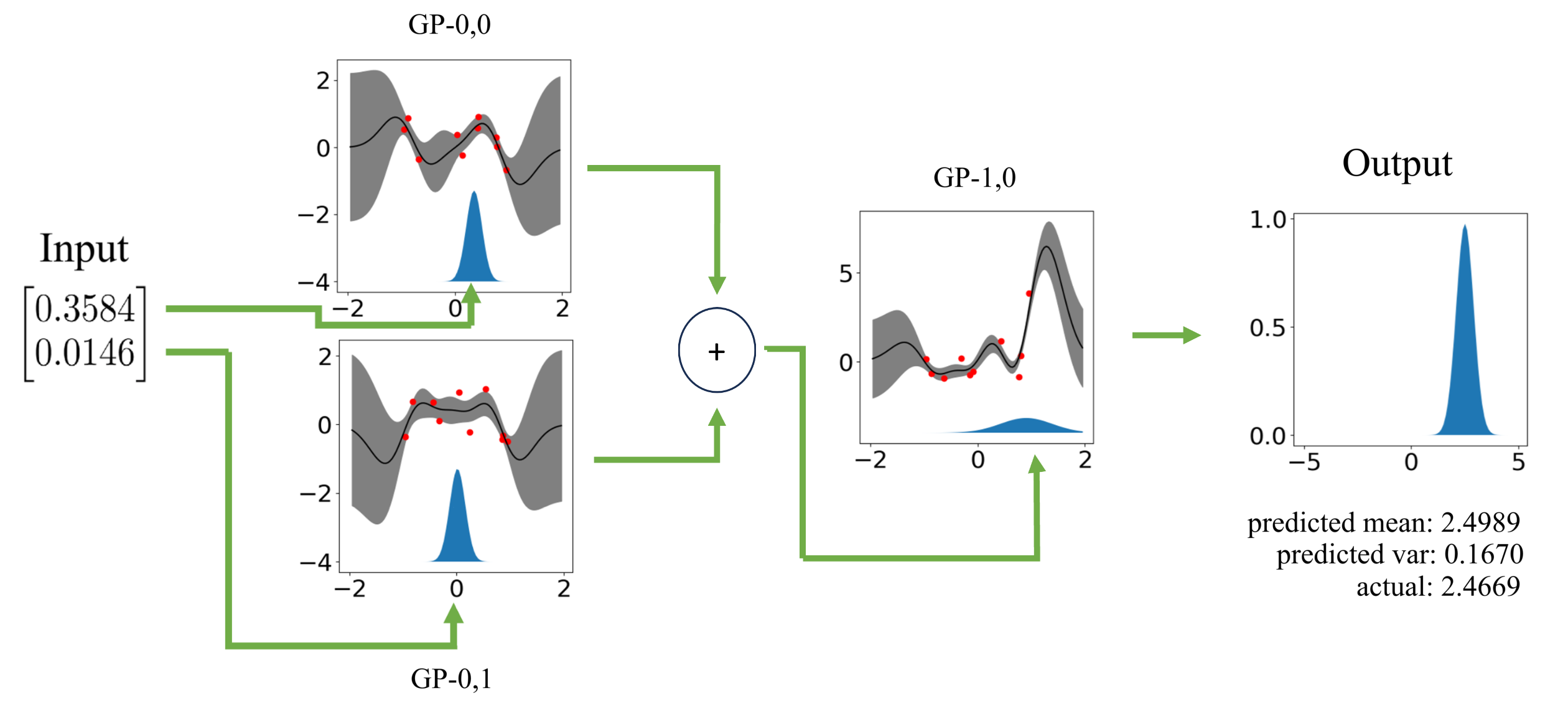}
        \caption{}
    \end{subfigure}
    \hfill
    \begin{subfigure}[b]{0.4\columnwidth}
        \includegraphics[width=\textwidth]{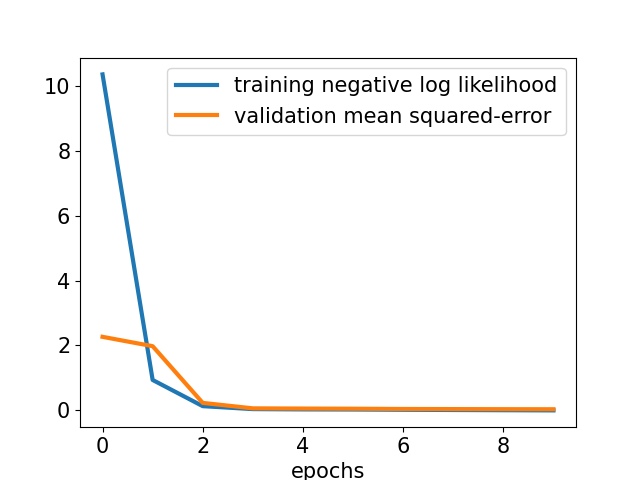}
        \caption{}
    \end{subfigure}

    \caption{(a) Experiment to model $y=\exp(\sin(\pi x_1)+x_2^2)$ using a 2-1 GP Neuron layout, where two GP Neurons consume $x_1$ and $x_2$ respectively, and their output is summed and consumed by another GP Neuron. Note that output of each GP Neuron is a Gaussian Distribution, and summation refers to summation of Gaussian-distributed variables following $\mathcal{N}(\mu_1,\sigma_1^2),\mathcal{N}(\mu_2,\sigma_2^2)$, giving $\mathcal{N}(\mu_1+\mu_2, \sigma_1^2 + \sigma_2^2)$. The red points represent the GP inducing points. (b) Training progress}
    \label{fig:simple_model}
\end{figure}

The GP neuron equivalent to a Fully-Connected layer can be defined, by using the binary operator of addition, as used in the Kolmogorov-Arnold Representation Theorem. Since each $\langle f, p(x)\rangle$ produces a Gaussian random variable, summation over several of them also produces a Gaussian random variable, leading to the expression below.
\begin{equation}
    \tilde{y}_i = \sum_{j=1}^{n} \langle \tilde{f}_{i,j}, p(x_j)\rangle, \quad \tilde{f}_{i,j}\sim\mathcal{GP}_{i,j}, \quad p(x_j)=\mathcal{N}(\mu_{x,j},\sigma_{x,j}^2)
\end{equation}
This can be generalised to a $m\times n$ matrix, where the operator $\ast$ is analogous to matrix multiplication, representing summation over elementwise inner product of functions, as described below.
\begin{equation}
    \begin{bmatrix}
        \tilde{y}_1 \\ \tilde{y}_2 \\ \vdots \\ \tilde{y}_m
    \end{bmatrix} =
    \begin{bmatrix}
        \tilde{f}_{1,1} & \tilde{f}_{1,2} & \hdots & \tilde{f}_{1,n} \\
        \tilde{f}_{2,1} & \tilde{f}_{2,2} & \hdots \\
        \vdots \\
        \tilde{f}_{m,1}
    \end{bmatrix}\ast
    \begin{bmatrix}
        p(x_1) \\ p(x_2) \\ \vdots \\ p(x_n)
    \end{bmatrix}
\end{equation}
Or more concisely:
\begin{equation}
    \begin{aligned}
        \tilde{Y}=\tilde{F}\ast p(X)
    \end{aligned}
\end{equation}

Each element of $\tilde{Y}$ is Gaussian-distributed and has an analytical form according to Equation \ref{eq:gpneuron}. The Fully-Connected GP layer takes in a vector of distributions $p(X)$ and produces a vector of distributions $p(Y)$. As noted in the previous section, the elements of $\tilde{Y}$ are only conditioned on the relevant GP inducing points and not on elements of $\tilde{X}$. Therfore, $\tilde{Y},\tilde{X}$ are independent Gaussians.

An experiment to model a simple 2-input function $y=\exp(\sin(\pi x_1)+x_2^2)$ was done, as shown in Figure \ref{fig:simple_model}, which is similar to the toy example in \cite{liu2024kan}. The training and validation datasets were chosen to have an input range of $[-0.5, 0.5]$, the GP inducing points were randomly initialised, and gradient-based optimization was performed. The model was able to quickly learn to minimise the mean square error between the prediction and validation labels. In Figure \ref{fig:simple_model}, the grey region represents 2 standard deviation from the mean for pointwise GP prediction. It can be observed that the GP-0,0 managed to learn a $\sin(\pi x_1)$ structure in the input range $[-0.5, 0.5]$. Similarly, the GP-0,1 can be interpreted to have learnt the $x_2^2$ structure for the input range $[-0.5, 0.5]$. GP-1,0 shows that it has learnt an approximate structure of $e^x$ although the structure appears less stable with some oscillatory features. Due to the flatter Gaussian distribution of the input to the 3rd GP neuron, the more fine-grained oscillatory feature of the 3rd GP neuron can be interpreted as having less influence on the neuron's output.

Significantly, each GP neuron appears to have learnt the exact univariate functions that compose the actual true function $y=\exp(\sin(\pi x_1)+x_2^2)$ using only random initialisation of inducing points. This shows the framework's potential in automatically uncovering and modelling the underlying non-linearities for a given problem.

\subsection{Convolutional GP Layer (ConvGP)}
\begin{figure}[t]
    \centering
    \includegraphics[width=0.9\columnwidth]{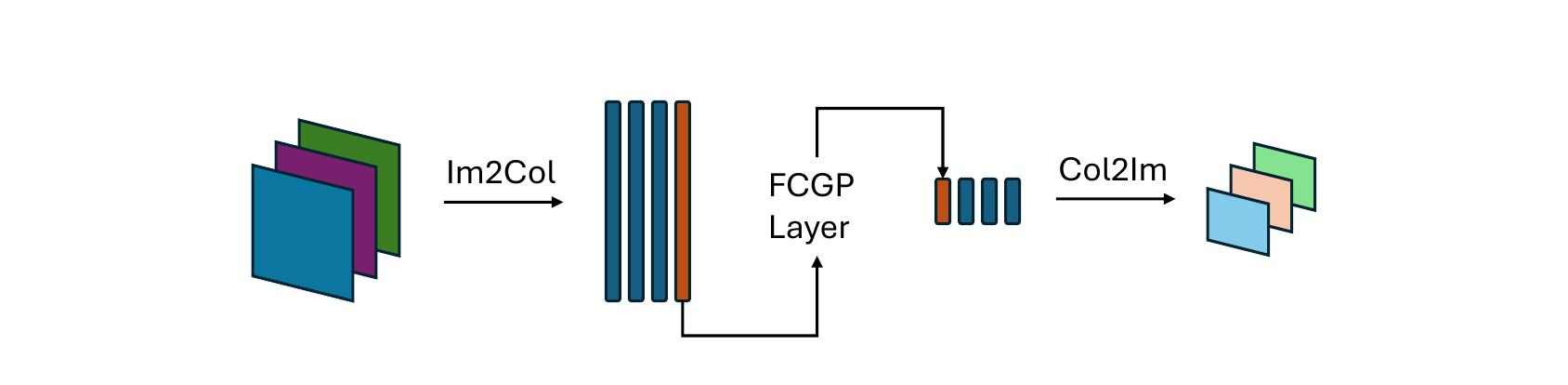}
    \caption{Convolutional GP (ConvGP) Layer}
    \label{fig:Convolutional_GP}
\end{figure}

Convolutional Neural Networks \cite{CNN} are a class of Neural Networks that are very effective in capturing translation-invariant features of input data and are central to solving image-related problems. One can define the GP neuron equivalent by transforming an image of pixels (where each pixel captures a Gaussian instead of a singular value) into vectors via the im2col routine \cite{im2col_first,im2col_phdthesis}. These vectors are then passed as a batch of input vectors through FCGP to output a batch of output vectors, which are then transformed back into an image (Figure \ref{fig:Convolutional_GP}) using col2im (inverse im2col). This is similar to the approaches in extending CNNs to KANs \cite{ConvolutionalKAN}. Since the same FCGP layer is used to process each input vector representing the moving convolution kernel and produces an output vector representing the elements across channels in a single pixel of the output image, each output pixel is considered \textit{conditionally independent} given the FCGP's internal GP reference points.

It is also noted that using 2 FCGP layers in series to form the moving kernel within a single ConvGP layer can improve modelling capabilities. This goes back to the Kolmogorov-Arnold Representation Theorem (Equation \ref{eq:Kolmogorov-Arnold-Theorem}), where equivalence of any multivariate function to a composition of univariate functions over the binary operation of addition requires the composition to have 2 layers.

\subsection{Activation Functions}
With the replacement of single values with Gaussians in the network, the selection of activation functions is now more restrictive if one aims to preserve the Gaussian-nature of the distributions. Linear functions such as AveragePool \cite{poolingmethods} are useable, since linear operations over Gaussians produce another Gaussian. However, the pooling window should be chosen to avoid overlapping that can introduce cross-correlation between output elements. Non-linear functions such as Sigmoid and Softmax do not preserve the Gaussian nature on input-output. In the case of Softmax, cross-correlation across the output elements are also introduced.

Instead, the FCGP layer is able to learn the non-linearity of activation functions such as Sigmoid and Softmax, while maintaining the independent Gaussian nature of the layer's output. This builds on the idea of learnable activation functions \cite{learnableActivations}, except that we use a GP here.

\subsection{Normalization}
It is beneficial to bound the distributions (mean and variance) of Gaussian r.v. in the graph, to avoid exploding gradients in back propagation during training. However, the set of Gaussian r.v. is only closed under linear operations, which does not serve well the purpose of bounding the distribution parameter values. Instead, a bijective mapping between Gaussian r.v. can be performed to keep the parameters well-behaved.

Given $\mathcal{G}$, denoting the set of unbounded Gaussian r.v. with mean $\mu\in(-\infty,\infty) $ and variance $\sigma^2\in[0,\infty)$, and $\mathcal{G}'$, the set of bounded Gaussian r.v. with mean $\mu\in[a,b]$ and variance $\sigma^2\in[0,c]$, where $a,b$ and $c$ are finite, a bijective map $\gamma: \mathcal{G} \rightarrow \mathcal{G}'$ can be chosen, such as the following:
\begin{equation}
    \begin{aligned}
        \text{for }&x\in\mathcal{G},\quad x\sim\mathcal{N}(\mu_x,\sigma_x^2) \\
        \text{define }&y\sim\mathcal{N}(\mu_y, \sigma_y^2) \\
        \text{where }&\mu_y = \text{tanh}(\mu_x) \\
                     &\sigma_y^2 = \text{sigmoid}(\sigma_x^2 - \mu_x^2) \\
        \text{then }&y\in\mathcal{G}' \text{ where }a=-1,\space b=c=1
    \end{aligned}
    \label{eq:normalizeMap}
\end{equation}
Such a bijective map has the below properties, avoiding the problem of exploding gradients and stabilising the training process:
\begin{enumerate}
    \item for $\mu_x\rightarrow\pm\infty$, one gets $\mu_y\rightarrow\pm 1$ and $\sigma_y^2\rightarrow 0$
    \item for $\sigma_x^2\rightarrow\infty$, one gets $\sigma_y^2\rightarrow 1$
\end{enumerate}

\section{Experimental verification based on Image Classification}

\begin{figure}[h]
    \centering
    \begin{subfigure}[b]{0.3\columnwidth}
        \includegraphics[width=\textwidth]{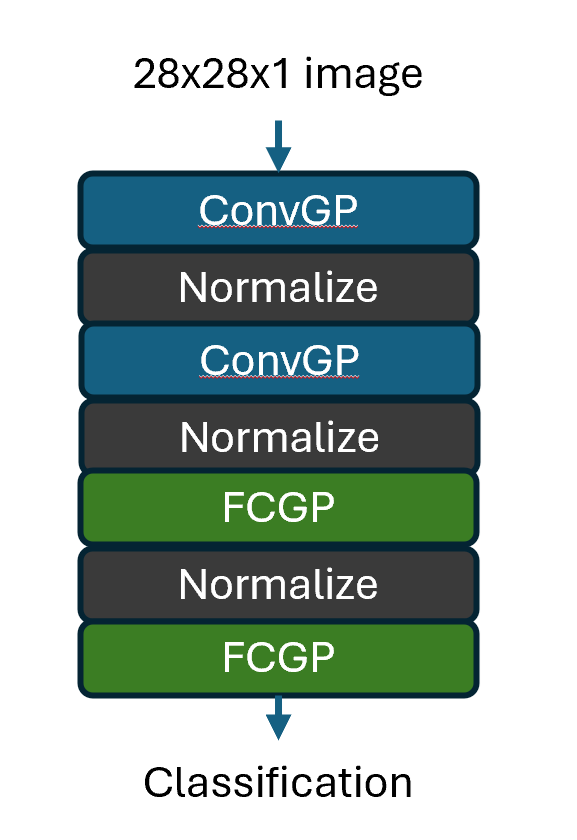}
        \caption{}
    \end{subfigure}
    \begin{subfigure}[b]{0.3\columnwidth}
        \includegraphics[width=\textwidth]{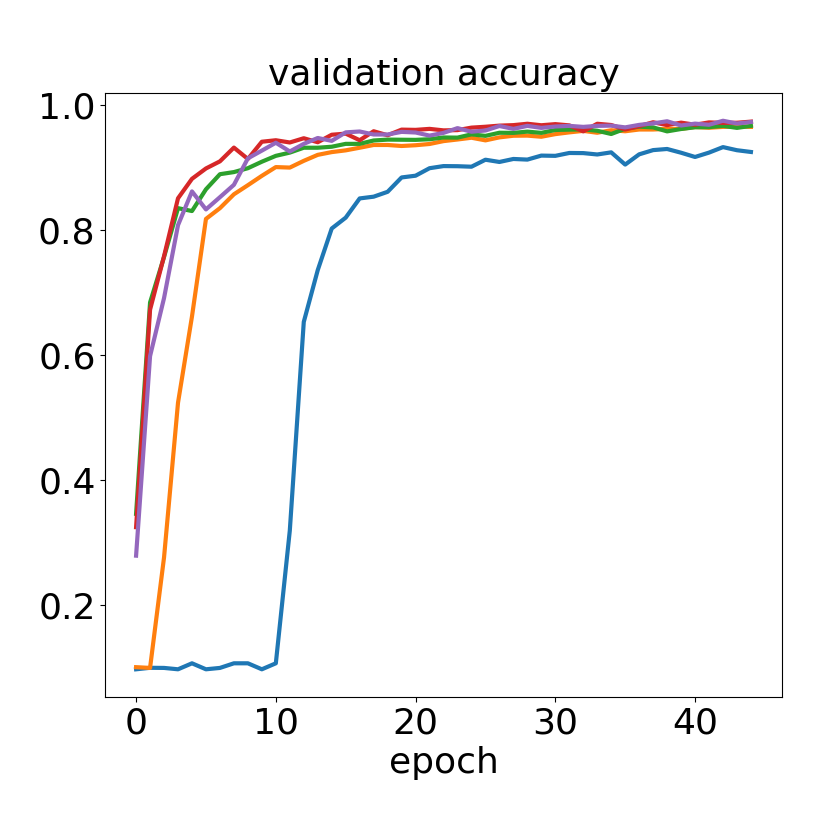}
        \caption{}
    \end{subfigure}
    \begin{subfigure}[b]{0.3\columnwidth}
        \includegraphics[width=\textwidth]{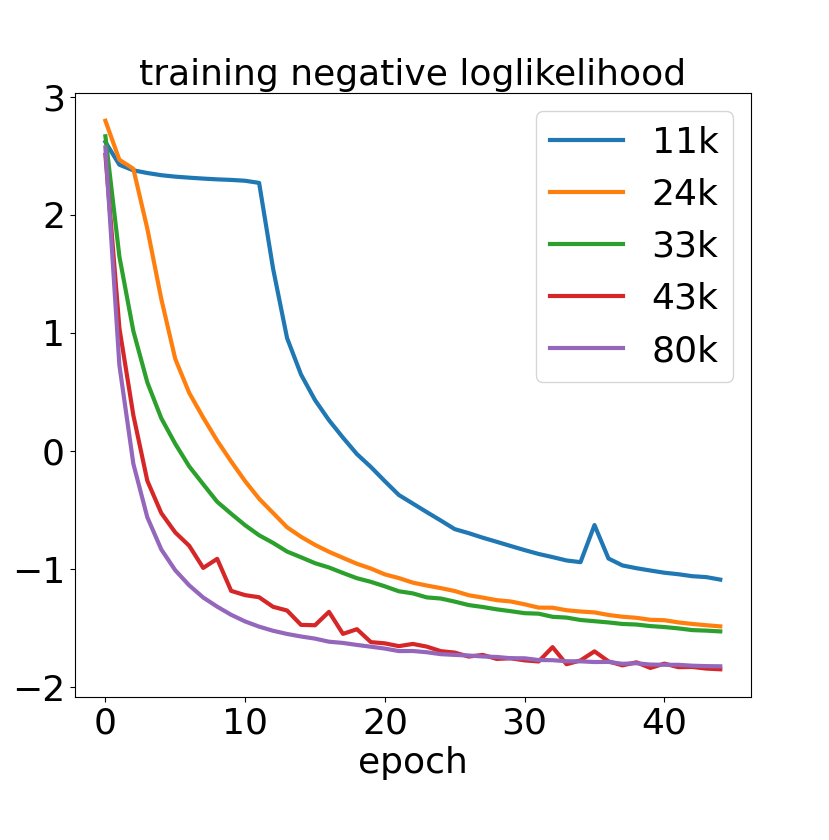}
        \caption{}
    \end{subfigure}
    \caption{(a) GP-KAN model structure used for MNIST classification. (b) Validation set accuracy and (c) Training negative log-likelihood over the course of training, for models of different parameter count, which is varied by changing the kernel, channel and stride values}
    \label{fig:MNIST_model}
\end{figure}

MNIST dataset \cite{MNIST} is a popular image classification dataset used to verify the validity of new model architectures, with current state-of-the-art \cite{NoRoutingModel} achieving 99.87\% at 1.5 million parameters. In Figure \ref{fig:MNIST_model}, the proposed GP-KAN network with a simple design  is trained on the MNIST dataset \cite{MNIST}. Every individual GP neuron is chosen to have 10 trainable inducing points. A train/validation/test split of 55k/5k/10k was used respectively, and the labels are transformed to one-hot encodings from 0-9. In the course of the training, it is observed that restricting $\sigma_n^2 > \lambda$ (Equation \ref{eq:GP_pred}) for a small chosen positive $\lambda$ helps to stablise the training and avoid the loss of Positive Semi-Definiteness of the covariance matrix. The variance of the input image is also treated as a learnable parameter.

\begin{figure}[t]
    \centering
    \begin{subfigure}[b]{0.7\columnwidth}
        \includegraphics[width=\textwidth]{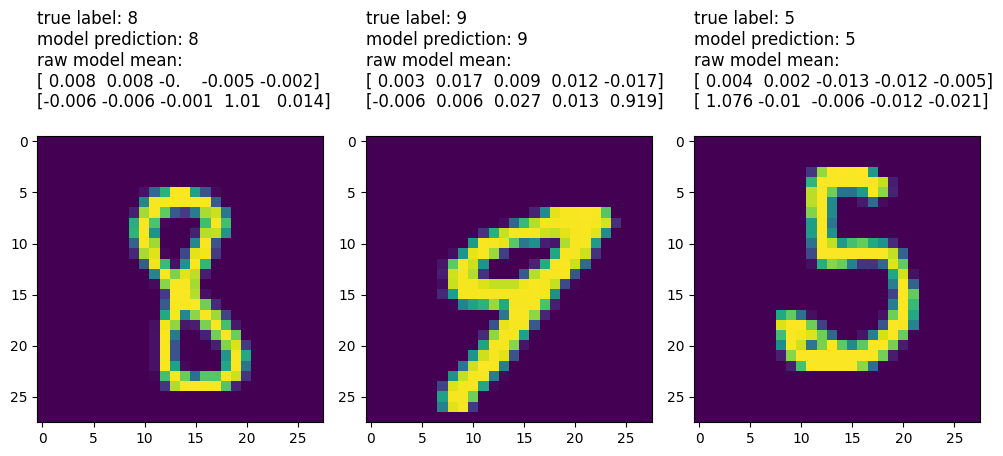}
        \caption{}
    \end{subfigure}
    \begin{subfigure}[b]{0.23\columnwidth}
        \includegraphics[width=\textwidth]{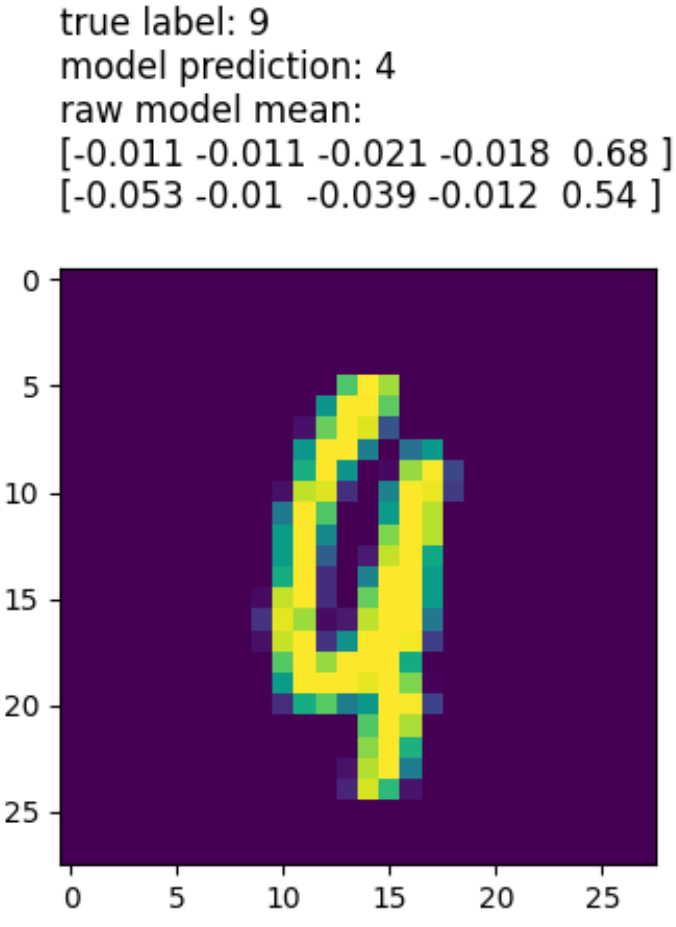}
        \caption{}
    \end{subfigure}
    \caption{Example of MNIST model predictions, with (a) showing the successful predictions and (b) showing a failed prediction. ArgMax is applied on the raw model output to get the index corresponding to the label. In the case of the failed prediction, the raw output values for the predicted index and the index corresponding to the true label are very close.}
    \label{fig:MNIST_example}
\end{figure}

The model was able to achieve good test accuracy of 98.53\% at 80 thousand parameters. Notably, even without the use of explicit activation functions (e.g. Softmax) designed for adapting model outputs to one-hot encodings, the model was able to automatically learn one-hot encoding style representation successfully (Figure \ref{fig:MNIST_example}). This is useful in situations where little is known about the nature of the desired model outputs beforehand.

\section{Conclusion}

In this paper, a Gaussian Process variant of the Kolmogorov Arnold Network architecture (GP-KAN) is proposed, where univariate GPs act as non-linear neurons. Unlike existing Deep GP models, a fully analytical uncertainty propagation method is used, along with various formulations of layers that can take advantage of this approach. GP-KAN has shown good modelling capabilities on the MNIST dataset with small number of parameters. Significantly, the absence of the need to choose explicit activation functions suggests that GP-KANs can potentially perform well on datasets where little to no prior knowledge of the nature of the outputs is available.

\bibliographystyle{hunsrt}
\bibliography{references}

\end{document}